\pdfoutput=1

\documentclass[11pt]{article}

\usepackage[preprint]{acl}

\usepackage{times}
\usepackage{latexsym}

\usepackage[T1]{fontenc}

\usepackage[utf8]{inputenc}

\usepackage{microtype}

\usepackage{inconsolata}

\usepackage{graphicx}
\usepackage{url}
\usepackage{hyperref}

\usepackage[frozencache,cachedir=.]{minted}

\title{Blocks Architecture (BloArk): Efficient, Cost-Effective, and Incremental Dataset Architecture for Wikipedia Revision History}

\author{
  \textbf{Lingxi Li\textsuperscript{1}} \;\;\;
  \textbf{Zonghai Yao\textsuperscript{1}} \;\;\;
  \textbf{Sunjae Kwon\textsuperscript{1}} \;\;\;
  \textbf{Hong Yu\textsuperscript{1,2,3,4}} \\
  \textsuperscript{1}Manning College of Information and Computer Sciences, University of Massachusetts Amherst \\
  \textsuperscript{2}Department of Medicine, University of Massachusetts Medical School \\
  \textsuperscript{3}Miner School of Computer and Information Sciences, University of Massachusetts Lowell \\
  \textsuperscript{4}Center for Healthcare Organization and Implementation Research, VA Bedford Health Care \\
  \texttt{\{lingxili,zonghaiyao,sunjaekwon\}@umass.edu, hong\_yu@uml.edu} \\
}

\begin{document}
\maketitle
\begin{abstract}
Wikipedia (Wiki) is one of the most widely used and publicly available resources for natural language processing (NLP) applications. Wikipedia Revision History (WikiRevHist)\footnote{\url{https://meta.wikimedia.org/wiki/Data_dumps}} shows the order in which edits were made to any Wiki page since its first modification. While the most up-to-date Wiki has been widely used as a training source, WikiRevHist can also be valuable resources for NLP applications. However, there are insufficient tools available to process WikiRevHist without having substantial computing resources, making additional customization, and spending extra time adapting others' works. Therefore, we report Blocks Architecture (BloArk), an efficiency-focused data processing architecture that reduces running time, computing resource requirements, and repeated works in processing WikiRevHist dataset. BloArk consists of three parts in its infrastructure: blocks, segments, and warehouses. On top of that, we build the core data processing pipeline: builder and modifier. The BloArk builder transforms the original WikiRevHist dataset from XML syntax into JSON Lines (JSONL) format for improving the concurrent and storage efficiency. The BloArk modifier takes previously-built warehouses to operate incremental modifications for improving the utilization of existing databases and reducing the cost of reusing others' works. In the end, BloArk can scale up easily in both processing Wikipedia Revision History and incrementally modifying existing dataset for downstream NLP use cases. The source code\footnote{GitHub: \url{https://github.com/lilingxi01/bloark}}, documentations\footnote{Documentations: \url{https://bloark.lingxi.li/}}, and example usages\footnote{Example usages: \url{https://wikidata.lingxi.li/}} are publicly available online and open-sourced under GPL-2.0 license.
\end{abstract}

\begin{figure*}[!t]
  \centering
  \includegraphics[width=0.8\linewidth]{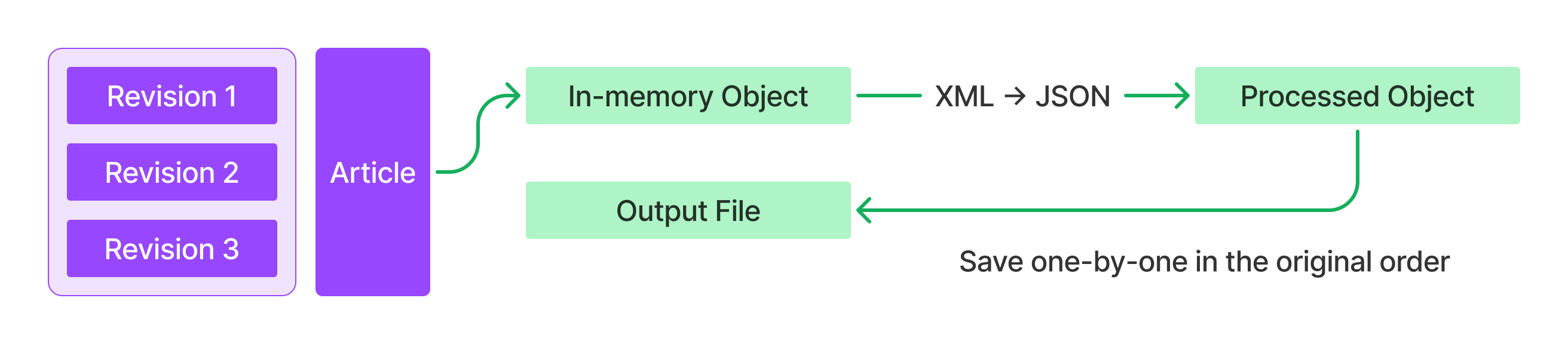}
  \caption{Traditional single-process multi-thread Python script for processing WikiRevHist. It needs to have the entire XML file decompressed into disk space before parsing, load revisions one-by-one, transform to JSON objects, apply changes, then store to JSON files.}
  \label{fig:traditional_arc}
\end{figure*}

\section{Introduction}

Wikipedia has played an important role in natural language processing (NLP) areas, such as information extraction \cite{kwon2022medjex, tran2014wikipevent, fisichella2021event, althoff2015timemachine, liu2020named}, rephrasing \cite{botha2018learning, martinez2024split}, and relationship graphs \cite{gonzalez2022leveraging, piscopo2017makes, schmelzeisen2021wikidated, pellissier2019learning}. As most researchers embrace informative large language models (LLMs) trained on the latest snapshot of Wikipedia \cite{naveed2023comprehensive}, the value of the WikiRevHist dataset has been underrated. WikiRevHist is valuable for its nature of human editing records, which roots the human reasoning on how to create and revise documents in decades. However, existing methods for pre-processing complex NLP data like Pandas \cite{thiebaut2011processing, pivarski2020nested} are either requiring complicated setup or incompatible with the scale of WikiRevHist. While researchers have a lot of concurrent approaches to do batch processing of Wikipedia XML data dumps \cite{thiebaut2011processing}, those approaches require complex configurations and an extensive amount of computing resources online \cite{rawat2019automatic}. In addition, popular Python data processing libraries like Pandas have difficulties working with nested data structures \cite{pivarski2020nested} and do not have multiprocessing out of the box, which takes time to setup and overcome hardware bottlenecks. And finally, none of the data processing libraries provide an easy way to handle large-sized dataset, such as checking unit data structures, extracting metadata for faster queries, and limiting maximum disk space usage. This often results in the overuse of shared disk space on computing clusters and the failure of processing jobs. Therefore, a high-performing, cost-effective, and convenient solution for handling downstream works on the WikiRevHist dataset becomes significant.

To address this, we propose Blocks Architecture (BloArk), a new dataset architecture designed for processing WikiRevHist and building downstream datasets conveniently. To the best of our knowledge, this work is the first data architecture for processing WikiRevHist in an efficient, cost-effective, and incremental way.

To improve the computing resource utilization, BloArk uses multiprocessing, which divides a dataset building task into unit processing items and applies onto CPU cores in parallel. Unlike traditional single-process Python scripts in processing WikiRevHist as Figure \ref{fig:traditional_arc}, BloArk improves the processing speed by distributing the load onto multiple independent workers. From our experiment, parsing 50 dump files that have a total compressed size of 90 GB from the WikiRevHist took 12 hours 43 min using an Apple M1 chip with only one process, while the same process took 5 hours 19 min with four processes. As the extracted size of the entire WikiRevHist dump is more than 30 TB, researchers will spend days waiting for the process without using frameworks like Hadoop. Therefore, the ability to do parallel processing is what we consider first when designing this architecture.

To improve querying speed and dataset structure clarity, BloArk embeds metadata along with each warehouse. For example, article title and tags are saved in metadata to help filtering based on related categories. The byte offsets for each article in associated warehouse file are also saved to bring article-level concurrency into data processing.

Furthermore, to reduce the cost of reusing processed datasets, BloArk introduces a standardized protocol for all datasets created by BloArk. Researchers and prospective users can save a significant amount of time spent on adapting the dataset format of others' works on WikiRevHist. Users can easily access a preview of the dataset structure and make incremental modifications without the need for additional customization.

\section{Similar Frameworks}

While large-scale data analytics frameworks like Hadoop and Spark are convenient to use, they do not offer end-to-end toolkit such as data structure preview and row-level modifier defined as a Python function. While BloArk is not powerful and extensible comparing with enterprise-level data frameworks, BloArk is straightforward out-of-box, and does not require any complicated setup to run.

Furthermore, researchers can benefit from both BloArk and Spark. While Spark does not natively support XML dumps, it does support JSONL formats. Researchers can use BloArk to convert XML to JSONL and modify the row-level data structure through a straightforward definition. Subsequently, researchers can analyze the data stored in BloArk warehouses utilizing Spark.

\begin{figure*}[!t]
  \centering
  \includegraphics[width=1.0\linewidth]{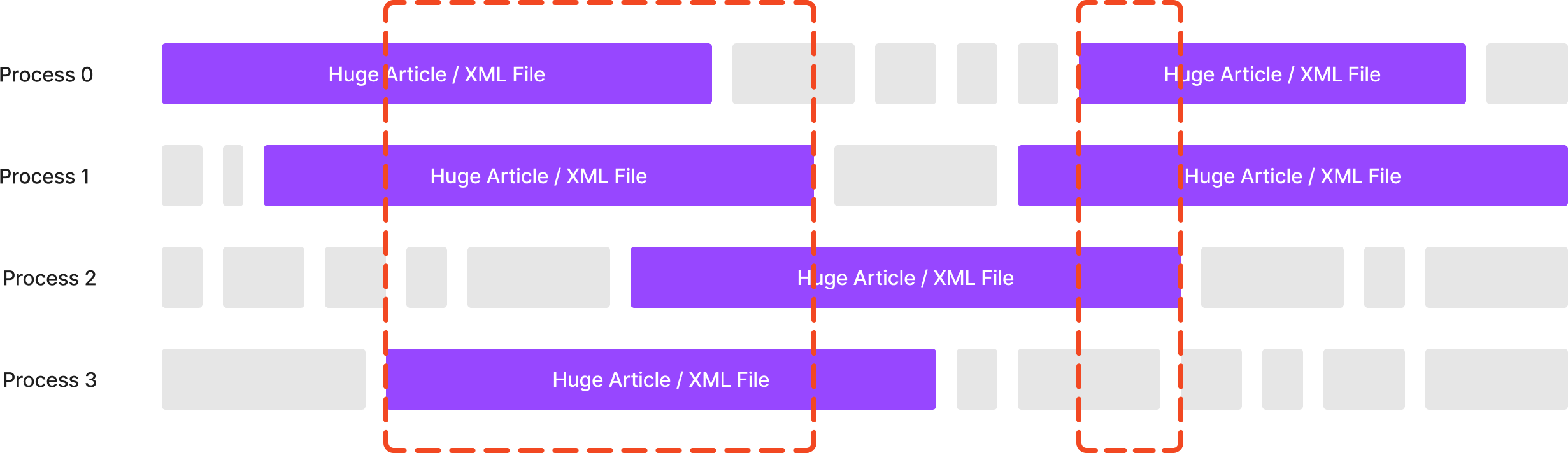}
  \caption{Understanding the congestion problem from a scheduling perspective. When we process huge articles or XML files at the same time, we also need to keep their decompressed files simultaneously, which increases the storage space bottleneck. Besides that, since large items take longer to process, disk space usage can easily accumulate because large articles are more likely to collide than smaller articles.}
  \label{fig:scheduling_problem}
\end{figure*}

\section{Background}

\subsection{File Format}

One of the most critical factors of efficiency is concurrency. We decided to use JSONL as the base file format for expanding the possibility of concurrency. JSONL is similar to JSON (JavaScript Object Notation) structure, which uses curly brackets for embracing an object with key-value pairs. JSON Lines (as known as JSONL format) have one JSON at a line, where the root of the file represents a list of objects. The benefit of using JSONL structure is that the processing of a file does not have to be linear. It is possible only to read the third object of a JSONL file without reading the first two objects. File formats like JSON and XML require linear parsing, which is not feasible for parallel processing within a file.

In general cases, the parallel processing will stay at the file level, such that one file would only be assigned to one process. In contrast, BloArk expands the parallelism to article-level. When transforming JSONL files, BloArk assigns the same JSONL file into multiple processes, where each process knows the starting and ending byte offset that corresponds to an article. In this way, only certain bytes of data is loaded for one process, which avoids memory overhead and I/O bottleneck.

\begin{figure*}[!t]
  \centering
  \includegraphics[width=1.0\linewidth]{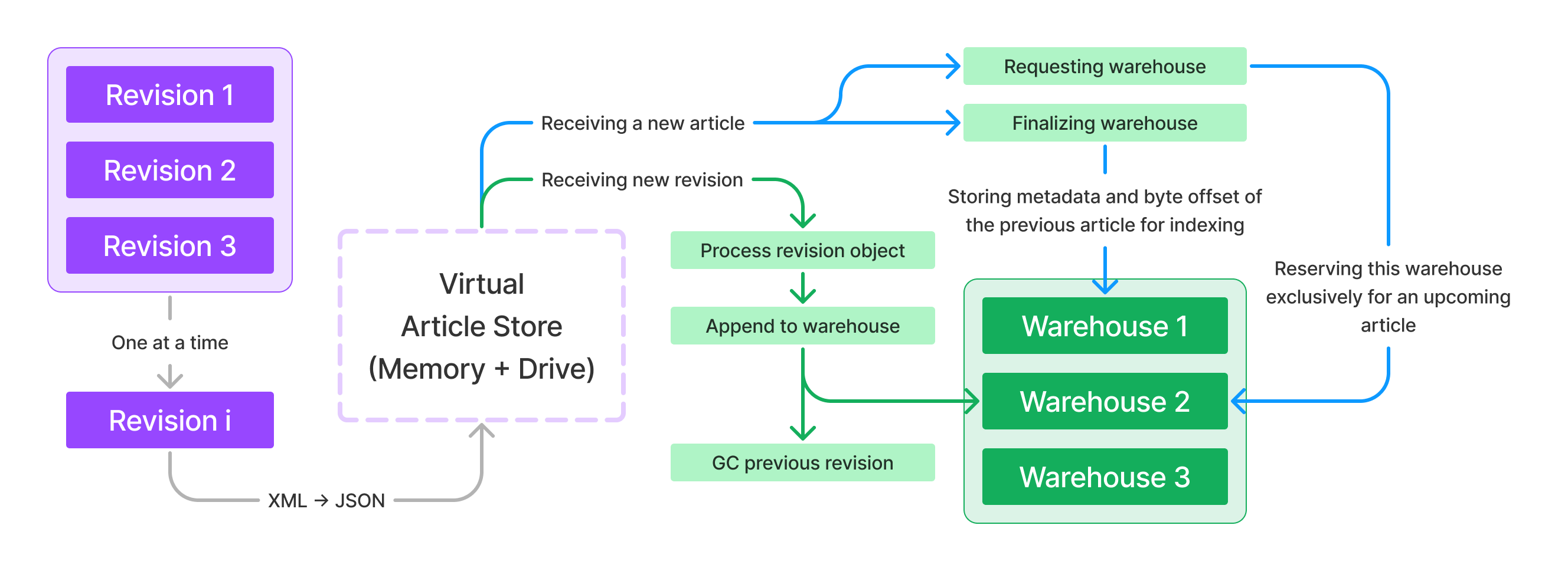}
  \caption{BloArk execution diagram of the "building process" in a single-process perspective. BloArk reads each XML file from top to bottom and at the third depth. When a revision is received, we store each revision as one JSON object in the warehouse file (JSONL format) and store the metadata in a separate, uncompressed JSONL file. When a new article is detected, we finalize the previous article and assign a new warehouse for storage.}
  \label{fig:final_arc}
\end{figure*}

\subsection{Unit Independence}
\label{sec:independence}

First, one XML file from the original WikiRevHist dataset contains many articles that are independent of each other, which could be sent to different processes for improving running efficiency. However, in the given flow, there are two steps that require a linear processing: XML reading and JSONL writing. In the raw WikiRevHist dataset, the structure of one XML file looks like this:

\begin{minted}[fontsize=\small]{xml}
<mediawiki>
  <siteinfo>...</siteinfo>
  <!-- First article -->
  <page>
    <title>...</title>
    <id>...</id>
    <!-- First revision -->
    <revision>
      <id>...</id>
      <parentid>...</parentid>
      <timestamp>...</timestamp>
      <text>...</text>
    </revision>
    <!-- Second revision -->
    <revision>
      ...
    </revision>
  </page>
  <!-- Second article -->
  <page>
    ...
  </page>
</mediawiki>
\end{minted}

XML needs to be read line-by-line because each object consists of a starting tag and an ending tag. Without finding the ending tag, we cannot finalize the current object and cannot start accepting the next object. Besides that, we need to read XML into objects at revision level instead of article level because some articles with a few thousands of revisions can easily exceed the memory limit. Therefore, within one XML file, this reading process is forced to be linear. Although there exists an approach \cite{zhang2022parallel} to parallelize the XML parsing process for speeding up, this approach requires chunking XML files and eventually loading the entire XML file into memory, which is not feasible for dataset having large individual files, such as WikiRevHist.

The same situation happens in writing JSONL file as well, where each line of a JSONL file is a complete JSON object. Even though we have independence between lines, we cannot write the next line efficiently until the current line is completed. There are also potential writing conflicts between processes without locking, so it is best practice to only allow one process to write a JSONL file.

Ultimately, most researchers end up utilizing only one CPU per XML file, which leads to a potential issue: the necessity for an excessive amount of storage space on shared clusters to accommodate decompressed XML files concurrently.

\subsection{Unit Processing Item and Resource Congestion}

The unit processing item, such as all revisions of one article, is an important factor in dealing with large-sized datasets like WikiRevHist. Loading all revisions of an article as a unit processing item can be oversized for the memory when the revision size is large, especially in a concurrent scenario where all processes share the same memory. For instance, some articles that have ~300K revisions could easily take up to 60 GB of memory when we are loading from a raw XML file, making changes, and outputting to a JSON file.

In addition, due to the nature of Python multiprocessing, no exception will be thrown from the child process if it runs out of memory, such as when all revisions of a long article are loaded into memory at once. The unhandling scenario like this increases the engineering complexity and failure rate especially on large datasets like WikiRevHist. In our experiments, resource congestion problem as described in Figure \ref{fig:scheduling_problem} can be observed frequently.

To resolve this, we define the unit processing item to be per revision instead of per article, which reduces the amount of data loaded into memory at the same time, and improves the concurrency. We will discuss this in detail in a later section.

\subsection{Reusability}

When conducting research using WikiRevHist, a significant challenge arises. Researchers often encounter the need for extensive additional work to adapt and utilize others' works effectively. Decompression and recompression processes were configured repeatedly, and dataset structures varied. To improve reusability and downstream research collaborations, BloArk standardizes the data structure for revision-based dataset such as WikiRevHist and embeds repeated works within a unified data pipeline. Any dataset built from BloArk should be effortlessly imported, viewed, and updated without the need of heavy engineering configurations. Details of our approach will be described in the "modifying process" below.

\begin{figure}[!t]
  \centering
  \includegraphics[width=0.9\linewidth]{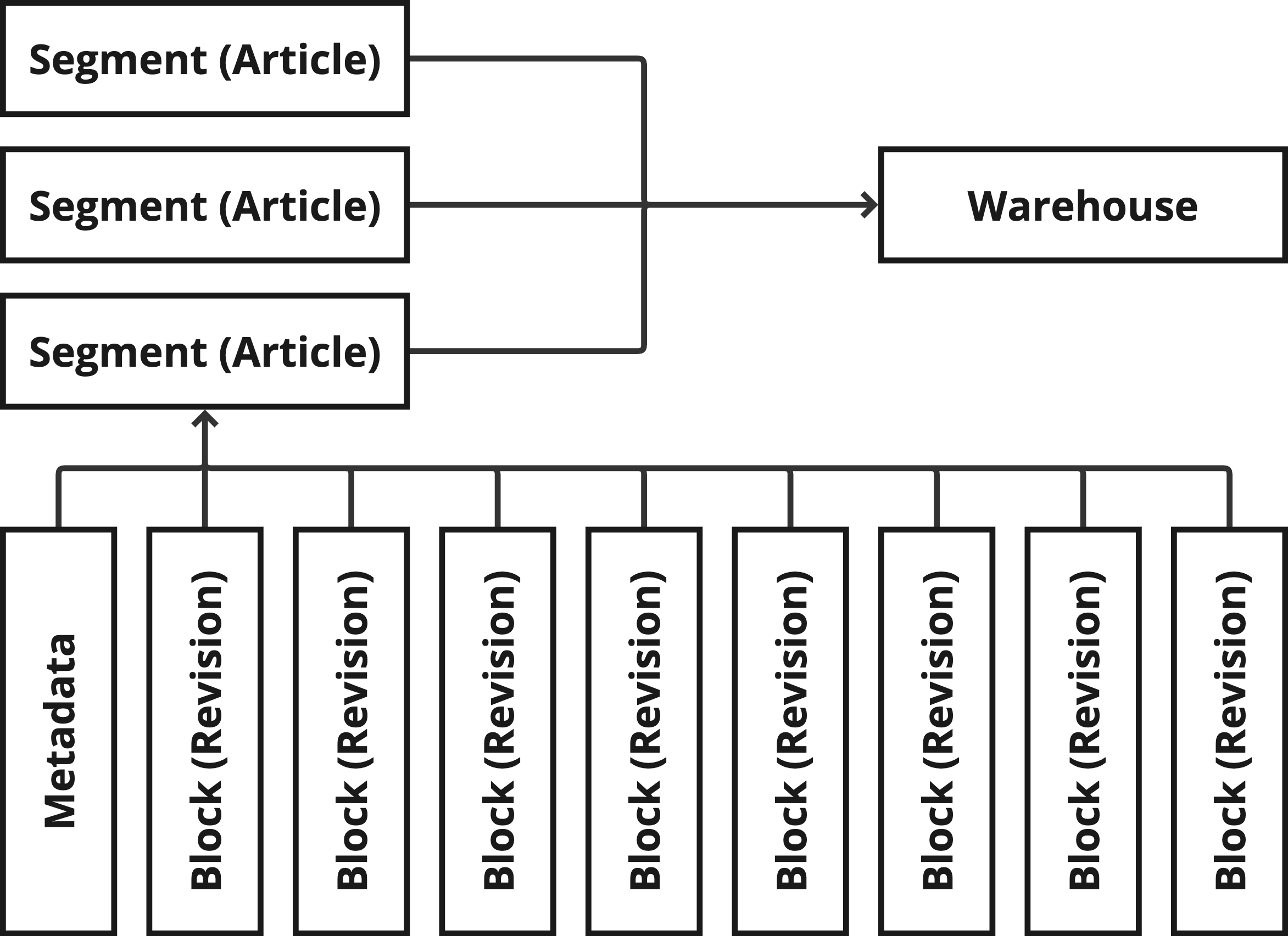}
  \caption{BloArk's data structure has three components: blocks, segments, and warehouses. In the mapping to WikiRevHist, a block represents a revision, a segment consists of a metadata object and all revisions on a timely basis, and a warehouse contains multiple segments (articles) until exceeding the size limit.}
  \label{fig:bloark_layout}
\end{figure}

\begin{figure*}[!t]
  \centering
  \includegraphics[width=1.0\linewidth]{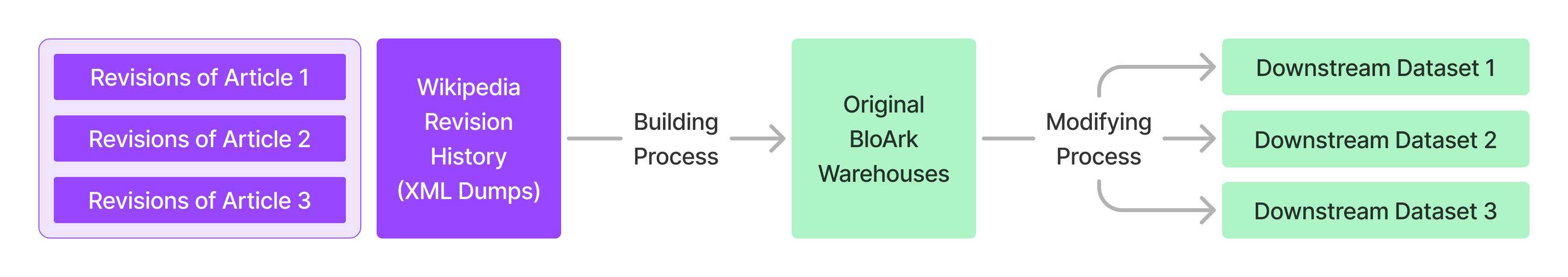}
  \caption{BloArk's data flow reduces the processing cost by making all downstream datasets on top of an original processed BloArk warehouses, which transforms XML data into JSON format. Under this setting, the most time-consuming process, the "building process", will only be executed once for an entire WikiRevHist dump of a month.}
  \label{fig:data_flow}
\end{figure*}

\section{Architecture and Usage}
\label{sec:architecture}

One goal of this research is to build a highly reusable architecture for supporting a wide range of downstream research in exploring the potential value of WikiRevHist. Since Wikipedia XML dumps are difficult to handle and expensive to process \cite{thiebaut2011processing}, BloArk transforms the XML dumps into JSONL format before any data processing for easier storage and handling. This is called the "building process". This process is used for overcoming the high cost of processing XML files by only executing it once. After the "building process", researchers can create downstream datasets based on a modifier that tells BloArk how to transform each unit processing item in the dataset. This is called the "modifying process".

\subsection{Building Process}
\label{sec:building-process}

In the building process, we use parallel CPU cores to decompress raw XML dumps of WikiRevHist and transform revision objects from XML syntax into JSON format before storing into the disk, as illustrated in Figure \ref{fig:final_arc}. Due to the independence reason mentioned in Section \ref{sec:independence}, the optimal approach is to process one XML file per CPU core, which makes this step harder to scale.

As described in Figure \ref{fig:bloark_layout}, BloArk consists of three parts: blocks, segments, and warehouses. These three components are shared across two processes. In the mapping to the WikiRevHist, a block corresponds to a single revision, a segment represents an article comprising multiple revisions, and a warehouse consists of a specified number of articles, defined by a size limit. Each block represents an JSON object equivalent to the structure below:

\begin{minted}[fontsize=\small]{json}
{
  "article_id": "...",
  "revision_id": "...",
  "timestamp": "...",
  "contributor": {
    "username": "...",
    "id": "..."
  },
  "comment": "...",
  "format": "...",
  "text": {
    "@bytes": "...",
    "#text": "..."
  },
  "sha1": "..."
}
\end{minted}

\subsection{Modifying Process}
\label{sec:modifying-process}

After building the raw dataset from XML dumps into BloArk warehouses, we can make batch changes to the existing dataset. To configure the modifying process, researchers need to define a BloArk modifier that takes revision information in each step, and outputs the target block that should be stored. Article-level computations can also be done through segment metadata, such as word counts and article URL extractions. This step is similar to MapReduce \cite{dean2008mapreduce}, which applies batch changes to the dataset, but BloArk modifier is simpler to define and easier to use on smaller-sized machines. We will describe the example usage and process setup in Section \ref{sec:example}. To avoid overflowing the memory in subprocesses, BloArk loads blocks only when it is requested, and discards the loaded variable once the modification of a block has been done. For example, if we are trying to extract link differences between adjacent revisions from WikiRevHist while discarding all irrelevant information, the modified block will be structured like this:

\begin{minted}[fontsize=\small]{json}
{
  "article_id": "...",
  "revision_id": "...",
  "timestamp": "...",
  "added_urls": ["...", "...", "..."],
  "removed_urls": ["...", "...", "..."]
}
\end{minted}

\subsection{Parallelization}

In order to deliver a similar performance as Hadoop while keeping usability and convenience on smaller machines, we specify the unit processing item for all BloArk jobs. Unit processing item is the minimum unit that its peer can be safely processed in other CPU cores without duplicating efforts. In the "building process", a unit processing item is one XML file. Even though articles are independent from each other on Wikipedia, they are stored in a way that has a linear dependency in the XML dumps. For example, we cannot get the second article in a certain way without going through the first article. After building the warehouses, BloArk stores the file offset for a segment, so it is fast and convenient to locate the revisions of an article without needing to go through the articles stored before it. Therefore, in the "modifying process", it is possible to process articles from the same warehouse across multiple CPU cores. This increases the computing resource utilization when processing size is small.

\section{Example Usages}
\label{sec:example}

In this section, we demonstrate the complete usage of BloArk library from downloading the source dataset, building original warehouses, to modifying previously-built warehouses based on specific research needs. The complete data flow for WikiRevHist downstream datasets is illustrated in Figure \ref{fig:data_flow}.

Please note that Python snippets in this section are simplified for demonstration purposes. They are designed to be run in Jupyter Notebooks. Additional code and type verification might be needed to run them directly as a Python script, such as:

\begin{minted}[fontsize=\small]{python}
if __name__ == '__main__':
    # Your code snippet goes to here
\end{minted}

\subsection{Download the Source Data}

Before building the original warehouses, the source WikiRevHist data dump is required, such as English Wikipedia (enwiki)\footnote{\url{https://dumps.wikimedia.org/enwiki/}} hosted on Wikimedia Foundation. It can be downloaded efficiently using WikiDL library\footnote{WikiDL Docs: \url{https://wikidl.lingxi.li/}} and with a maximum of 3 processes for a fair use of public resources\footnote{This 3-process limit is observed from Wikimedia gateway rules. HTTP Error 503 will be returned if having more than 3 parallel downloads.}. The code sample for downloading WikiRevHist is demonstrated below. Downloading may require a significant amount of time.

\begin{minted}[fontsize=\small]{python}
from wikidl import WikiDL

downloader = WikiDL(
    # Specify parallel downloading (max 3).
    num_proc=3,
    # Update this to the latest dump date.
    snapshot_date='20240801',
    # This means: Edit History Dump (EHD).
    select_pattern='ehd',
)
# Process starts.
downloaded_files = downloader.start(
    # Save all compressed dumps into `/input`.
    output_dir='./input',
)
\end{minted}

\subsection{Build Original Warehouses from the Source Data}

The "building process" of BloArk should be applied to transform original XML dumps of WikiRevHist dataset into BloArk warehouses in JSONL format. As described in Section \ref{sec:architecture}, the "building process" is required for any downstream dataset and expected to only run once.

For better system reliability, it is recommended to reserve at least 1 GB of memory per CPU in this long-running job. This memory limit depends on the largest size of an article. Memory overflow is generally difficult to identify in Python, and it leads to a CPU process that never joins back to the main process. As the WikiRevHist is updated every month, larger memory budget per CPU is recommended to avoid losing long-running progress.

\begin{minted}[fontsize=\small]{python}
import bloark

builder = bloark.Builder(
    # Define the output location for warehouses.
    output_dir='./warehouses',
    # Use 8 processes (CPUs) in parallel.
    num_proc=8,
)

# Load all compressed XML dump file names.
# It does not load files into memory yet.
builder.preload('./input')

# Optional: if you want to test with the first 10
# compressed XML dumps, use following line.
# builder.files = builder.files[:10]

# This command will take a long time.
builder.build()
\end{minted}

\subsection{Example Dataset: Clean Text and Links}

All WikiRevHist contents use Wikitext, a markup language for all Wikipedia documents. To extract clean text that does not include any markup syntax for better downstream training, we propose a new dataset that can be easily built using BloArk. In the "modifying process", we define the block-level modifier function using Grimm package\footnote{Grimm Package Docs: \url{https://twiki.lingxi.li/docs/grimm/get-started}} and store texts, links, and images as new blocks into new warehouses.

Cleaned WikiRevHist data has been widely used in training editing models, such as in modeling editing processes \cite{reid2022learning} task. BloArk can improve the efficiency of data preparation by simplifying the implementation and enhancing the processing speed.

\begin{minted}[fontsize=\small]{python}
import bloark
from grimm import clean_syntax

class CleanModifier(bloark.ModifierProfile):
    def block(
        self, content: dict, metadata: dict
    ):
        text_content = content['text']['#text']
        output = clean_syntax(text_content)
        text, ext_urls, int_urls, imgs = output

        new_content = {
            "revision_id": content['revision_id']
            "clean_text": text,
            "external_links": ext_urls,
            "internal_links": int_urls,
            "images": imgs,
        }
        return new_content, metadata

modifier = bloark.Modifier(
    output_dir='./output',
    num_proc=8,
)

# Load original warehouses.
modifier.preload('./warehouses')

# Tips: you can add more than one profile.
modifier.add_profile(CleanModifier())

# This command will take a long time.
modifier.start()
\end{minted}

The original input of this process shapes as described in Section \ref{sec:building-process}. After running the "modifying process", outputted blocks in new warehouses will be structured like this:

\begin{minted}[fontsize=\small]{json}
{
  "revision_id": "...",
  "clean_text": "...",
  "external_links": ["...", "...", "..."],
  "internal_links": ["...", "...", "..."],
  "images": ["...", "...", "..."]
}
\end{minted}

\subsection{Example Dataset: 6-Month Snapshots}
\label{sec:snapshot-dataset}

One way to modify original warehouses is by filtering, such as keeping only revisions that meet specific criteria. This can also help reduce the size of the dataset and the cost of future processing. In the past, significant efforts have focused on generating the next revision of an article based on previous revision histories. In those NLP tasks, WikiRevHist can conveniently provide article snapshots every six months within the past decade. Therefore, we propose a new dataset based on BloArk that has revision snapshots of an article for every 6 month. The block-level structure of this dataset should remain the same as described in Section \ref{sec:building-process}, but have less blocks.

There are two benefits. First, it is easier to observe apparent changes in snapshots every 6 months than continuous editing histories. When using all editing revisions to train the model, some revisions might not help generalize the pattern of changes for the actual event, as those are simply replacing some unnecessary words or fine-tuning paragraphs. Second, WikiRevHist data hosting platforms like Wikimedia Foundation does not keep latest revision data dumps that are older than 3 months, which makes it very hard to find the snapshot at a specific time frame from the internet without accessing the full revision history.

The following is a simplified example code for modifying this dataset from original warehouses.

\begin{minted}[fontsize=\small]{python}
from datetime import datetime, timedelta
import bloark

class SnapshotModifier(bloark.ModifierProfile):
    last_date: datetime = None

    def block(
        self, content: dict, metadata: dict
    ):
        timestamp = content['timestamp']
        curr_date = datetime \
            .fromisoformat(timestamp)

        if self.last_date and curr_date < (
            self.last_date + timedelta(days=180)
        ):
            # Return `None` to not save this block.
            # `metadata` is still needed.
            return None, metadata
        
        self.last_date = curr_date
        return content, metadata

modifier = bloark.Modifier(
    output_dir='./output',
    num_proc=8,
)
modifier.preload('./warehouses')
modifier.add_profile(SnapshotModifier())
modifier.start()
\end{minted}

\subsection{Example Dataset: 6-Month Edit Summaries}

In the task of summarizing human edits, we need a dataset that contains the edit differences and a generated summary on those differences. Original WikiRevHist dump kept all revisions, which is too frequent for this dataset and is not cost-effective to have a large amount of generation works. Therefore, we propose a new dataset based on BloArk to extract the summary of edits from each article in a 6-month time frame. For every adjacent revisions in an article, we compare their text, get a list of differences, and use LLM to generate a summary.

With the incremental modification by BloArk, the creation process of this dataset could be based on the 6-month snapshot dataset mentioned in Section \ref{sec:snapshot-dataset}. It saves time on re-filtering revisions from the source, and it is convenient to reuse works that had already been done by BloArk.

\begin{minted}[fontsize=\small]{python}
import bloark

class SummaryModifier(bloark.ModifierProfile):
    last_text: str = None

    def block(
        self, content: dict, metadata: dict
    ):
        if not last_text:
            last_text = curr_text
            return None, metadata

        curr_text = content['text']['#text']

        # TODO: Implement this function.
        changes = diff_function(...)

        # TODO: Implement this function.
        summary = summarize_changes(...)

        last_text = curr_text
        new_data = {
            "changes": changes,
            "summary": summary,
            "timestamp": content['timestamp'],
        }
        return new_data, metadata

modifier = bloark.Modifier(
    output_dir='./output',
    num_proc=8,
)

# Load the previously-built snapshot dataset.
# This saves the time to filter from source.
modifier.preload('./6-month-snapshots')
modifier.add_profile(SummaryModifier())
modifier.start()
\end{minted}

The modified block will have differences for every 6 month, and be structured as below. The actual format of edit differences will be based on the implementation of difference function.

\begin{minted}[fontsize=\small]{json}
{
  "changes": [
    { "type": "add", "content": "..." },
    { "type": "remove", "content": "..." },
  ],
  "summary": "...",
  "timestamp": "..."
}
\end{minted}

\section{Limitations and Future Works}

First, current BloArk does not have a way to incrementally sync changes when the source XML dumps are updated. WikiRevHist dumps update once a month. Therefore, users need to rebuild from the source every month in order to get the most up-to-date dataset. In the future, the "building process" of BloArk can be expanded with a feature to extract differences between two XML dumps and update previously-built warehouses from the differences.

Second, raw WikiRevHist XML dumps store each revision in full text. To improve storage efficiency, users can extract differences between adjacent revisions using libraries like difflib or ergodiff, and only store the differences. This extraction process can be achieved with a BloArk modifier applied after the "building process".

Third, current BloArk does not support the separation of blocks. This can be improved by designing a new API for modifiers, which allows returning multiple blocks instead of requiring one-on-one mapping. This future work can be widely used on tasks like expanding a single revision into multiple knowledge entries where each block is a tuple for knowledge graph.

Lastly, this work currently lacks a benchmark or evaluation. Establishing an empirical benchmark to evaluate the efficiency of data processing frameworks on WikiRevHist would be beneficial for comparing the performance and usability among similar frameworks. Additionally, it would serve as a measuring guideline for future research in this area.

\section{Conclusion}

In this work, we introduce BloArk, an efficient, cost-effective, and incremental dataset architecture for processing WikiRevHist. BloArk provides two different processes, the "building process" and the "modifying process", for resolving two main issues: high cost of handling XML dumps, and inconvenience of querying and modifying existing datasets built upon XML dumps. Since all datasets built by BloArk can be easily imported and modified further, the cost of doing research on WikiRevHist will be decreased. With BloArk, prospective users can save their time when exploring the potential value of WikiRevHist and other downstream datasets.

\bibliography{bloark}

\appendix

\section{Distribution and Maintenance}
\label{sec:faq}

\begin{itemize}
    \item \textbf{Will the source code of BloArk be open sourced on public platforms? Will it be published?} \\ Yes. BloArk is open sourced on GitHub under GPL-2.0 license. Everyone is welcomed to submit issues/pull requests (PRs) on BloArk's GitHub public repository. BloArk package is published on PyPI and free to download for everyone using Python package manager.
    \item \textbf{When will the source code be distributed?} \\ The source code is immediately available on our GitHub public repository.
    \item \textbf{Who will be supporting/maintaining the BloArk?} \\ Lingxi Li will maintain the BloArk code base on GitHub and publish version changes to PyPI periodically. Bug reports can be opened on GitHub issues and Lingxi Li will address them by severity.
    \item \textbf{How can the owner/curator/manager of the dataset architecture be contacted (e.g., email address)?} \\ Lingxi Li, the creator/maintainer of BloArk, can be contacted at: \textit{research@lingxi.li}.
    \item \textbf{Will downstream datasets be distributed publicly?} \\ No. BloArk is a data processing architecture that can be used to build datasets. It is not a dataset. Downstream datasets will be built, distributed, and owned by prospective users.
    \item \textbf{Will original warehouses be distributed to third parties outside of the entity (e.g., company, institution, organization) on behalf of which the dataset was created?} \\ Yes, but for sample access only. Users can use BloArk package and example code provided above to replicate the "building process" and build original warehouses on their own resources. We may consider releasing one version of original warehouses built from WikiRevHist XML dumps to Hugging Face for public sample access.
    \item \textbf{Is there an erratum?} \\ BloArk has changelogs recorded in its official website\footnote{\url{https://bloark.lingxi.li/resources/changelog}}. This information will also be available on GitHub publishes.
    \item \textbf{Will BloArk be updated (e.g., bug fixes, performance improvements, feature requests)?} \\ Lingxi Li will fix severe bugs and monitor GitHub issues for bug reports and questions. Feature requests and performance improvements will be made by maintainers' decisions. Since BloArk is open sourced, everyone can contribute to the code base, and Lingxi Li will review the contribution to ensure the quality and safety of BloArk.
    \item \textbf{Has BloArk been used for any tasks already?} \\ BloArk has already been used in tasks given in example datasets described in Section \ref{sec:example}.
    \item \textbf{Will older versions of BloArk continue to be hosted?} \\ All previous versions of BloArk package will always be available to download through Python package manager from PyPI.
    \item \textbf{If others want to extend/augment/build on/contribute to this dataset architecture, is there a mechanism for them to do so?} \\ Yes. BloArk's GitHub repository is public and opened to everyone for contributions through PR. Lingxi Li will review submitted code to ensure quality and safety of BloArk package.
    \item \textbf{Will BloArk be distributed under a copyright or other intellectual property (IP) license, and/or under applicable terms of use (ToU)?} \\ BloArk is open sourced under GPL-2.0 license. The copyright of WikiRevHist dataset belongs to its original license from Wikipedia. All downstream datasets will not have ownership connection to BloArk.
\end{itemize}

\end{document}